\documentclass[letterpaper, 10 pt, conference]{ieeeconf}  

\IEEEoverridecommandlockouts                              

\usepackage{graphicx} 
\usepackage{amsmath} 
\usepackage{amssymb}  
\usepackage{algorithm}
\usepackage{algpseudocode}
\usepackage{mathtools}
\usepackage{multirow}
\usepackage[caption=false]{subfig}
\usepackage[dvipsnames]{xcolor}
\usepackage{soul}

\usepackage{xparse}

\title{\LARGE \bf
Learning Functionally Decomposed Hierarchies for Continuous Control Tasks with Path Planning
}

\author{Sammy Christen$^{1}$, Lukas Jendele$^{1}$, Emre Aksan$^{1}$ and Otmar Hilliges$^{1}$
\thanks{$^{1}$AIT Lab, Department of Computer Science, ETH Zurich, 8092
Zurich, Switzerland
        {\tt\small sammy.christen@inf.ethz.ch}}%
}

\ExplSyntaxOn
\NewDocumentCommand{\fpcompare}{mmm}
 {
  \fp_compare:nTF { #1 } { #2 } { #3 }
 }
\ExplSyntaxOff
\makeatletter
\def\mathcolor#1#{\@mathcolor{#1}}
\def\@mathcolor#1#2#3{%
  \protect\leavevmode
  \begingroup
    \color#1{#2}#3%
  \endgroup
}
\makeatother

\newif\ifshowcomments
\showcommentstrue 

\ifshowcomments
        \newcommand{\note}[3]{{\color{#2}[#1: #3]}}
        \newcommand{\OH}[1]{\note{OH}{red}{#1}}
		\newcommand{\EA}[1]{\note{EA}{blue}{#1}}
		\newcommand{\LJ}[1]{\note{LJ}{green}{#1}}
		\newcommand{\SC}[1]{\note{SC}{blue}{#1}}
		
\else
        \newcommand{\note}[3]{\unskip}
        \newcommand{\OH}[1]{\unskip}
		\newcommand{\EA}[1]{\unskip}
		\newcommand{\LJ}[1]{\unskip}
		\newcommand{\SC}[1]{\unskip}
\fi

\makeatletter
\newcommand\footnoteref[1]{\protected@xdef\@thefnmark{\ref{#1}}\@footnotemark}
\makeatother

\DeclareMathOperator*{\argmax}{arg\,max}

\makeatletter
\let\NAT@parse\undefined
\makeatother
\usepackage{hyperref}

\begin{document}

\maketitle
\thispagestyle{empty}
\pagestyle{empty}

\newcommand*\squeezespaces[1]{
  \thickmuskip=\scalemuskip{\thickmuskip}{#1}%
  \medmuskip=\scalemuskip{\medmuskip}{#1}%
  \thinmuskip=\scalemuskip{\thinmuskip}{#1}%
  \nulldelimiterspace=#1\nulldelimiterspace
  \scriptspace=#1\scriptspace
}
\newcommand*\scalemuskip[2]{%
  \muexpr #1*\numexpr\dimexpr#2pt\relax\relax/65536\relax
} 

\begin{abstract}
\setcounter{footnote}{1}
We present HiDe, a novel hierarchical reinforcement learning architecture that successfully solves long horizon control tasks and generalizes to unseen test scenarios. Functional decomposition between planning and low-level control is achieved by explicitly separating the state-action spaces across the hierarchy, which allows the integration of task-relevant knowledge per layer. We propose an RL-based planner to efficiently leverage the information in the planning layer of the hierarchy, while the control layer learns a goal-conditioned control policy. The hierarchy is trained jointly but allows for the modular transfer of policy layers across hierarchies of different agents. We experimentally show that our method generalizes across unseen test environments and can scale to 3x horizon length compared to both learning and non-learning based methods. We evaluate on complex continuous control tasks with sparse rewards, including navigation and robot manipulation.

\end{abstract}

\section{Introduction}

Reinforcement learning (RL) can solve long horizon control tasks with continuous state-action spaces in robotics \cite{lillicrap2015ddpg, levine2015deepvisuo, schulman2017proximal}, such as robot manipulation \cite{vulin2021}  or human-robot interaction \cite{christen2019}. However, tasks that involve extended planning and sparse rewards still pose many challenges in successfully reasoning over long horizons and in achieving generalization from training to different test environments. Therefore, hierarchical reinforcement learning (HRL) splits the decision making problem into several subtasks at different levels of abstraction \cite{sutton1999tempabstr, andre2002funcabstrac}, often learned separately via curriculum learning \cite{frans2017meathierarchy, bacon2016optioncritic, sasha2017fun}, or end-to-end via off-policy and goal-conditioning \cite{levy2018hierarchical, nachum2018data, nachum2018nearoptimal}. However, these methods share the full-state space across layers, even if low-level control states are not strictly required for planning. This limits i) modularity in the sense of transferring higher level policies across different control agents and ii) the ability to generalize to unseen test tasks without retraining.

\begin{figure}[t!]
	\centering
	\includegraphics[width=0.95\columnwidth]{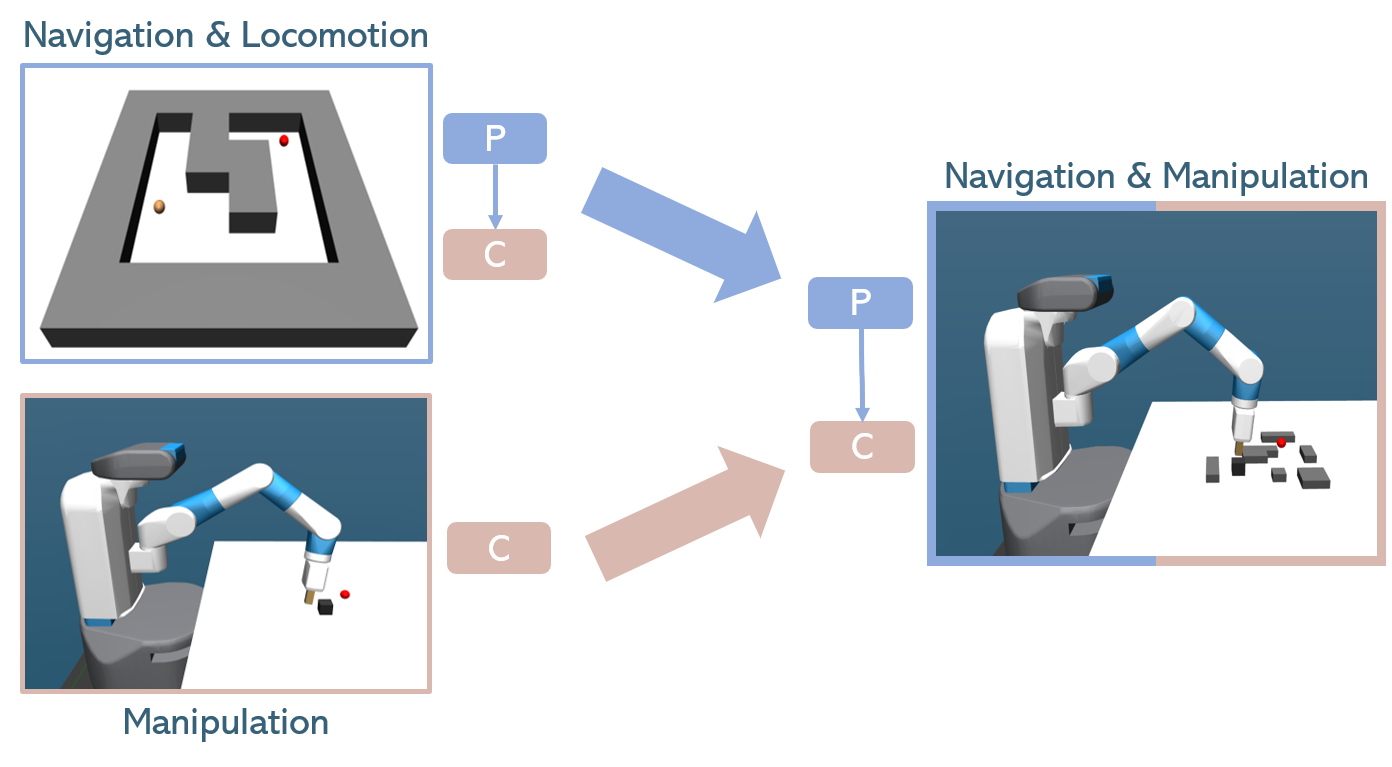}
	\vspace*{-5mm}
	\caption{Our functionally decomposed hierarchical architecture allows training higher-level \textbf{P}lanning policies with simple agents and combining them with more complex \textbf{C}ontrol policies. Example transfer of the planning of a simple 2DoF ball agent trained on a locomotion task to a manipulation robot to solve more complex tasks such as pushing a cube around obstacles.}
	\label{fig:teaser}
	\vspace{-0.4cm}
\end{figure}   

In this paper, we study how a more explicit hierarchical task decomposition into local control and global planning tasks can alleviate both issues. In particular, we hypothesize that explicit decoupling of the state-action spaces of different layers, as is done in robotics \cite{mueller2018}, can be beneficial for end-to-end learned hierarchies. More sepcifically, it allows the integration of suitable inductive biases that can be leveraged to achieve generalization across agents and environments. Thus, we propose a 2-level hierarchy (see Fig. \ref{fig:hierarchy}) that is suited for continuous control tasks with a 2D-planning component. Furthermore, we show in our experiments that planning in 3D is also possible.
Global environment information is only available to the planning layer, whereas the full internal state of the agent is only accessible by the control layer. To leverage the global information, we propose the integration of an efficient, RL-based planner.  

The benefit of this explicit task decomposition is manyfold. First, the individual layers have access only to task-relevant information, enabling layers to focus on their individual tasks \cite{dayan2000feudalrl}.  Second, the modularity allows for the composition of new agents without retraining. We demonstrate this via transferring the planning layer across different low-level agents ranging from a simple 2DoF ball to a 17DoF humanoid. The approach even allows to generalize across domains, combining layers from navigation and robotic manipulation tasks to solve a compound task (see Fig. \ref{fig:teaser}).

In our framework, which we call HiDe, a goal-conditioned control policy $\pi_{\text{c}}$ on the lower-level of the hierarchy interacts with the environment. It has access to the proprioceptive state of an agent and learns to achieve subgoals $g$ that are provided by the planning layer policy $\pi_{\text{p}}$. The planning layer has access to task-relevant information, e.g., an occupancy grid, and needs to find a subgoal-path towards a goal. We stress that the integration of such additional information into HRL approaches is non-trivial. For example, naively adding an image to HRL methods \cite{levy2018hierarchical, nachum2018data} causes an explosion of the state-space complexity and hence leads to failure as we show in Section \ref{sec:maze_nav}. We propose a specialized, efficient planning layer, based on MVProp \cite{nardelli2018value} with an added learned dynamic agent-centric attention window which transforms the task-relevant prior into a value map. The action of $\pi_{\text{p}}$ is the position that maximizes the masked value map and is fed as a subgoal to the control policy $\pi_{\text{c}}$. While the policies are functionally decoupled, they are trained jointly, which we show to be beneficial over separately training a control agent and attaching a conventional planner.

We focus on continuous control problems that involve navigation  and path planning from top-down view, e.g., an agent navigating a warehouse or a robot arm pushing a block. However, we show as a proof of concept that HiDe can also work in non-Euclidean space and plan in 3D. In our experiments, we first demonstrate that generalization and scaling remain challenging for state-of-the-art HRL approaches and are outperformed by our method. We also compare against a baseline with a non-learning based planner, where a control policy trained with RL is guided by a conventional RRT planner \cite{rrt}. We then show that our method can scale to 3x longer horizons and generalize to randomly configured layouts. Furthermore, we demonstrate transfer across agents and domains. 
The results indicate that an explicit decomposition of policy layers in combination with task-relevant knowledge and an efficient planner are an effective tool to help generalize to unseen environments and make HRL more practicable. In summary our main contributions include:

\begin{itemize}
    \item A novel HRL architecture that enforces functional decomposition into global planning and low-level control through a strict separation of the state space per layer, combined with an RL-based planner on the higher layer of the hierarchy, to solve long horizon control tasks.
    \item We provide empirical evidence that task-relevant priors are essential components to enable generalization to unseen test environments and to scale to larger environments, which HRL methods struggle with.
    \item Demonstration of transfer of individual modular layers across different agents and domains.
\end{itemize}
\vspace{-0.2cm}
\section{Related Work}
\label{sec:rw}

\subsubsection{Hierarchical Reinforcement Learning}
Learning hierarchical policies has seen lasting interest \cite{sutton1999tempabstr,dayan2000feudalrl, schmidhuber1991learningTG,dietterich1991maxq}, but many approaches are limited to discrete domains. Sasha et. al \cite{sasha2017fun} introduce FeUdal Networks (FUN), inspired by \cite{dayan2000feudalrl}. In FUN, a hierarchic decomposition is achieved via a learned state representation in latent space, but only works with discrete actions.
More recently, off-policy methods that work for goal-conditioned continuous control tasks have been introduced  \cite{levy2018hierarchical, nachum2018data, nachum2018nearoptimal, tirumala2019, abbeel_scoop}.
Nachum et. al \cite{nachum2018data, nachum2018nearoptimal} present HIRO and HIRO-LR, an off-policy HRL method with two levels of hierarchy. The non-stationary signal of the upper policy is mitigated via off-policy corrections. In HIRO-LR, the method is extended by learning a representation of the state and subgoal space space from environment images. In contrast to our approach, both methods use a dense reward function. Levy et. al \cite{levy2018hierarchical} introduce Hierarchical Actor-Critic (HAC) that can jointly learn multiple policies in parallel via different hindsight techniques from sparse rewards. HAC, HIRO and HIRO-LR consist of a set of nested policies where the goal of a policy is provided by the top layer. In contrast to our method, the same state space is used in all layers, which prohibits transfer of layers across agents. Similar to \cite{abbeel_scoop,konidaris2007}, we introduce a modular design to decouple the functionality of individual layers. This allows us to define different state, action and goal spaces for each layer. In contrast to \cite{abbeel_scoop}, our method can be trained end-to-end and scales to longer horizons.
Our method is closest to HIRO-LR, which also has access to a top-down view map. Although the learned space representation of HIRO-LR can generalize to a mirrored environment, the policies need to be retrained for each task. Contrarily, HiDe generalizes without retraining through the explicit use of the map for planning.

\subsubsection{Planning in Reinforcement Learning}
In model-based RL, much attention has been given to learning of a dynamics model of the environment and subsequent planning \cite{sutton1990ntegratedAF, wang2019modelbased}. Eysenbach et. al \cite{eysenbach2019search} propose a planning method that performs a graph search over the replay buffer. However, they require to spawn the agent at different locations in the environment and let it learn a distance function in order to build the search graph. Unlike model-based RL, we do not learn state transitions explicitly. Instead, we learn a spatial value map from collected rewards.
Recently, differentiable planning modules that are trained via model-free RL have been proposed \cite{nardelli2018value, tamar2016value, oh2017vpn, srinivas2018upn}. Tamar et. al \cite{tamar2016value} establish a connection between CNNs and Value Iteration \cite{bertsekas2000dpoc}. They propose \emph{Value Iteration Networks} (VIN), where model-free RL policies are additionally conditioned on a fully differrentiable planning module. MVProp \cite{nardelli2018value} extends this by making it more parameter-efficient and generalizable.
Our planning layer is based on MVProp. However, we do not rely on selecting an action from a fixed neighborhood mask (see Section \ref{sec4_method}). Instead we learn an attention mask which is used to generate intermediate goals for the low-level policy. We also extend our planner to 3D, whereas MVProp is only shown in 2D.
Gupta et. al \cite{gupta2017cognitive} learn a map of indoor spaces and do planning using a multi-scale VIN. However, the robot operates only on a discrete set of macro actions. Nasiriany et. al \cite{nasiriany2019planning} use a goal-conditioned policy for learning a TDM-based planner on latent representations.  Srinivas et. al \cite{srinivas2018upn} propose Universal Planning Networks (UPN), which also learn how to plan an optimal action trajectory via a latent space representation. Müller et. al. \cite{mueller2018} separate planning from low-level control to achieve generalization by using a supervised planner and a PID-controller.
In contrast to our approach, the latter methods either rely on expert demonstrations or need to be retrained in order to achieve transfer to harder tasks.

\section{Background}

\subsection{Goal-Conditioned Reinforcement Learning}

\newcommand{\StatesSet}{\mathcal{S}}
\newcommand{\ActionsSet}{\mathcal{A}}
\newcommand{\TransitionSet}{\mathcal{T}}
\newcommand{\GoalsSet}{\mathcal{G}}
\newcommand{\state}[1]{{s}_{#1}}
\newcommand{\goal}[1]{{g}_{#1}}
\newcommand{\action}[1]{{a}_{#1}}
\newcommand{\reward}{r}
\newcommand{\discountRate}{\gamma}
\newcommand{\QFunction}{Q}
\newcommand{\policy}{\pi}
\newcommand{\Cost}{J}
\newcommand{\mdp}{\mathcal{M}}

We model a Markov Decision Process (MDP) augmented with a set of goals $\GoalsSet$. We define the MDP as a tuple $\mdp = \{\StatesSet, \ActionsSet, \GoalsSet, \mathcal{R},  \discountRate, \TransitionSet, \rho_0, \}$, where $\StatesSet$ and $\ActionsSet$ are set of states and actions, respectively, $\mathcal{R}_t=r(s_t,a_t, g_t)$ a reward function, $\gamma$ a discount factor $\in [0,1]$, $\TransitionSet = p(\state{t+1}|\state{t}, \action{t})$ the transition dynamics of the environment and $\rho_0 = p(\state{1})$ the initial state distribution, with $\state{t} \in \StatesSet$ and $\action{t} \in \ActionsSet$.  Each episode is initialized with a goal $g \in \mathcal{G}$ and an initial state is sampled from $\rho_0$. We aim to find a policy $\pi: \StatesSet \times \GoalsSet \rightarrow \ActionsSet$, which maximizes the expected return. We use an actor-critic framework where the goal augmented action-value function is defined as: $
Q(s_t,g_t, a_t) = \mathbb{E}_{a_t \sim \pi, s_{t+1} \sim \mathcal{T}} \left [\sum_{i=t}^{T}\discountRate^{i-t}\mathcal{R}_t \right ].$
The Q-function (critic) and the policy $\pi$ (actor) are approximated by using neural networks with parameters $\theta^Q$ and $\theta^{\policy}$. The objective for $\theta^Q$ minimizes the loss:
\begin{equation}
\begin{split}
&L(\theta^Q) = \mathbb{E}_{\mdp} \left [ \left( Q(\state{t}, \goal{t}, \action{t} ; \theta^Q) - y_t \right)^2   \right], \text{where}\\
&y_t = \reward(\state{t}, \goal{t}, \action{t}) + \discountRate Q(\state{t+1}, \goal{t+1}, \action{t+1} ; \theta^Q).
\label{eq:q_bellman}
\end{split}
\end{equation} 
The policy parameters $\theta^{\policy}$ are trained to maximize the Q-value:
\medmuskip=0mu
\thinmuskip=0mu
\thickmuskip=0mu
\begin{equation}
L(\theta^{\policy}) = \mathbb{E}_{\pi} \left [Q(\state{t}, \goal{t}, \action{t} ; \theta^Q)    \vert \state{t}, \goal{t}, \action{t} = \policy(\state{t}, \goal{t}; \theta^{\policy}) \right ]
\label{eq:pi}
\end{equation}

\subsection{Hindsight Techniques}
\label{sec3_hac}

In HAC, Levy et. al \cite{levy2018hierarchical} apply two hindsight techniques to address the challenges introduced by the non-stationary nature of hierarchical policies and the environments with sparse rewards. In order to train a policy $\pi_i$, optimal behavior of the lower-level policy is simulated by \emph{hindsight action transitions}. More specifically, the action $a_{i}$ of the upper policy is replaced with a state $s_{i-1}$ that is actually achieved by the lower-level policy $\pi_{i-1}$. Identically to HER \cite{andrychowicz2017hindsight}, \emph{hindsight goal transitions}
replace the subgoal $g_{i-1}$ with an achieved state $s_{i-1}$, which consequently assigns a reward to the lower-level policy $\pi_{i-1}$ for achieving the virtual subgoal. Additionally, a third technique called \emph{subgoal testing} is proposed. The incentive of subgoal testing is to help a higher-level policy understand the current capability of a lower-level policy and to learn Q-values for subgoal actions that are out of reach. Hence, a transition with a penalty reward is added to the replay buffer if the current lower-level policy $\pi_{i-1}$ cannot reach the provided subgoal within $H$ attempts. We find all three techniques effective and apply them to our model during training.

\subsection{Value Propagation Networks}

Tamar et. al \cite{tamar2016value} introduce value iteration networks (VIN) for path planning problems. Nardelli et. al \cite{nardelli2018value} propose value propagation networks (MVProp) with better sample efficiency and generalization behavior. MVProp creates reward- and propagation maps covering the environment. A reward map highlights the goal location and a propagation map determines the propagation factor of values through a particular location. 
The reward map is an image $\bar{r}_{i,j}$ of the same size as the environment image, where $\bar{r}_{i, j} = 0$ if the pixel $(i,j)$ overlaps with the goal position and $-1$ otherwise. The value map $V$ is calculated by unrolling max-pooling operations in a neighborhood $N$ for $k$ steps as follows: 
\begin{equation}
\centering
\begin{split}
v_{i,j}^{(0)} &= \bar{r}_{i,j} \\
\bar{v}_{i,j}^{(k)} &= \max_{(i',j')\in N(i,j)} \left(\bar{r}_{i,j}+p_{i,j}(v_{i',j'}^{(k-1)} - \bar{r}_{i,j})\right)\\
v_{i,j}^{(k)} &= \max\left (v_{i,j}^{(k-1)}, \bar{v}_{i,j}^{(k)} \right )
\end{split}
    \label{eq:mvprop_main}
\end{equation}
The action (i.e., the target position) is selected to be the pixels $(i', j')$ maximizing the value in a predefined $3x3$ neighborhood $N(i_0,j_0)$ of the agent's current position $(i_0,j_0)$:
\begin{equation}
    \pi(s,(i_0,j_0))=\underset{i',j'\in N(i_0,j_0)}{\argmax} v_{i',j'}^{(k)}
    \label{eq:mvprop_final}
\end{equation}

\begin{figure}[t]
	\centering
	\includegraphics[width=\columnwidth]{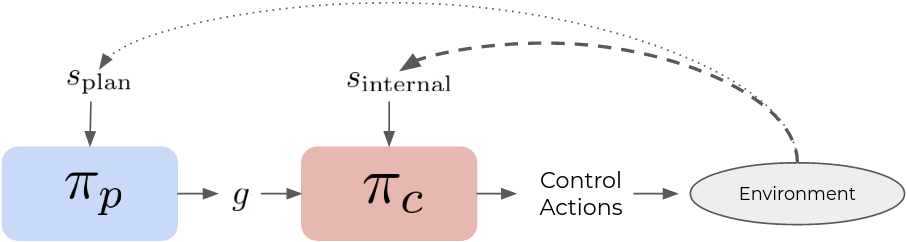}	
	\caption{Our HRL architecture. The planning layer policy $\pi_{\text{p}}$ receives global environment information $s_{\text{plan}}$ for planning and outputs a goal $g$ for the lower layer. The low-level control learns a goal-conditioned control policy $\pi_{\text{c}}$ from proprioceptive state information $s_{\text{internal}}$ to reach the provided goal.}
	\label{fig:hierarchy}
	\vspace{-0.3cm}
\end{figure}   

\section{Hierarchical Decompositional Reinforcement Learning}
\label{sec4_method}

\begin{figure*}[t]
    \vspace{0.2cm}
	\centering
	\includegraphics[width=1.5\columnwidth]{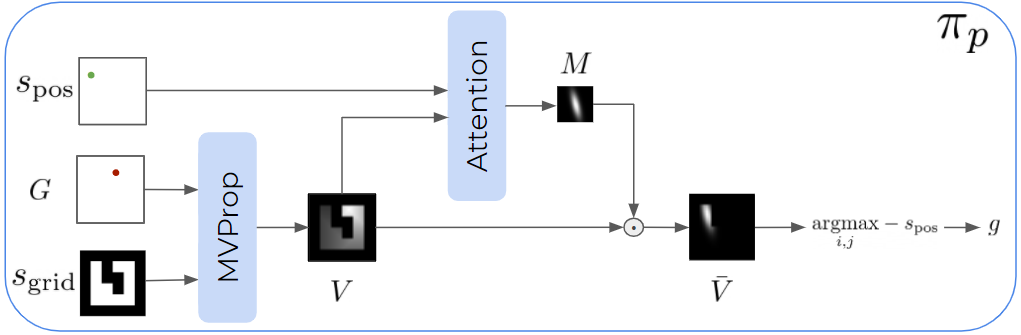}
	\caption{Planning layer $\pi_{\text{p}}(s_{\text{pos}}, s_{\text{grid}}, G)$ illustrated for the 2D case. Given an occupancy grid $s_{\text{grid}}$ and goal $G$, the MVProp network computes a value map $V$. An attention mask $M$, using the agent's position $s_{\text{pos}}$ restricts $V$ to a local subgoal map $\bar{V}$. The policy $\pi_{\text{p}}$ selects max value and assigns the control policy $\pi_{\text{c}}$ with a sugboal $g$ relative to the agent's position. For the 3D case, see Equations \ref{eq:gaussian_likelihood}-\ref{eq:masked_v_argmax} and Fig. \ref{fig:experiment_envs}c.}
	\label{fig:model_planner}
	\vspace{-0.5cm}
\end{figure*}

We introduce a 2-layer hierarchical architecture, HiDe, which allows for an explicit functional \emph{decomposition} across layers (see Fig. \ref{fig:hierarchy}). Our method achieves temporal abstractions via nested policies. Moreover, our architecture enforces functional decomposition explicitly by reducing the state in each layer to only task-relevant information. The planning layer is responsible for planning a path towards a goal and hence receives global information about the environment. The control layer has access to the agent's internal state and learns a goal-conditioned control policy that can achieve subgoals from the planning layer. The layers are jointly-trained by using the hindsight techniques and subgoal testing presented in Section \ref{sec3_hac} to overcome the sparsity of the reward and the non-stationarity caused by off-policy training. Our design significantly improves generalization and makes cross-agent transfer possible (see Section \ref{sec5_experiments}).

\subsection{Planning Layer}
\label{sec4_planninglayer}
The  planning layer is expected to learn high-level actions over a long horizon, which define a coarse path towards a goal. In related work \cite{levy2018hierarchical,nachum2018data,nachum2018nearoptimal}, the planning layer learns an \emph{implicit} value function and shares the same architecture as the lower layers. Since the task is learned for a specific environment, generalization is inherently limited. In contrast, we introduce a planning specific layer consisting of several components to learn the map and to find a feasible path to the goal. 

Our planning layer for the 2D case is illustrated in Fig. \ref{fig:model_planner}. We utilize a value propagation network (MVProp) \cite{nardelli2018value} to learn an \emph{explicit} value map which projects the collected rewards onto the environment space. For example, given a discretized 3D model of the environment, a convolutional network determines the per voxel flow probability $p_{i,j, k}$. The probability value of a voxel corresponding to an obstacle should be $0$ and that for free passages $1$, respectively. 

Nardelli et. all \cite{nardelli2018value} only use 2D images and a predefined $3\times3$ neighborhood of the agent's current position and pass the location of the maximum value in this neighbourhood as goal position to the agent (Equation \ref{eq:mvprop_final}). We extend this to be able to handle both 3D and 2D inputs and augment the MVProp network with an attention model which learns to define the neighborhood dynamically and adaptively. Given the value map $V$ and the agent's current position $s_{\text{pos}}$, we estimate how far the agent can move, modeled by a Gaussian. More specifically, we predict a full covariance matrix $\Sigma$ with the agent's global position $s_{\text{pos}}$ as mean. We later build a mask $M$ of the same size as the environment space $s_{\text{grid}}$ by using the likelihood function:
\begin{equation}
    m_{i,j,k} = \mathcal{N}((i,j,k) | s_{\text{pos}}, \Sigma)
    \label{eq:gaussian_likelihood}
\end{equation}

Intuitively, the mask defines the density for the agent's success rate. Our planning policy selects an action (i.e., subgoal) that maximizes the masked value map as follows:
\begin{equation}
    \begin{split}
        \bar{V} &= M \cdot V \\
        \pi_{\text{p}}(s_{\text{pos}}, s_{\text{grid}}, G) &= \underset{i,j,k}{\text{argmax  }} \bar{v}_{i,j,k} - s_{\text{pos}}\\
    \end{split}
    \label{eq:masked_v_argmax}
\end{equation}

where $\bar{v}_{i,j,k}$ corresponds to the value at voxel $(i,j,k)$ on the masked value map $\bar{V}$. Note that the subgoal selected by the planner is relative to the agent's current position $s_{\text{pos}}$, resulting in better generalization as we show in Section \ref{sec:ablation_study}. While we present the more general 3D case in Equations \ref{eq:gaussian_likelihood}-\ref{eq:masked_v_argmax}, we reduce the equations by one dimension for the 2D case used in most of our experiments.

The benefits of having an attention model are twofold. First, the planning layer considers the agent dynamics in assigning subgoals which may lead to fine- or coarse-grained subgoals depending on the underlying agent's performance. Second, the Gaussian window allows us to define a dynamic set of actions for the planning policy $\pi_{\text{p}}$, which is essential to find a path of subgoals on the map. While the action space includes all pixels of the value map $V$, it is limited to the subset of only reachable pixels by the Gaussian mask $M$. We find that this leads to better obstacle avoidance behaviour such as the corners and walls shown in Fig. \ref{fig:dynamic_static}.

Since our planning layer operates in discrete action space, the resolution of the projected image defines the minimum amount of displacement for the agent, affecting maneuverability. This could be tackled by using a soft-argmax \cite{chapelle2010gradient} to select the subgoal pixel, allowing to choose real-valued actions and providing invariance to image resolution. In our experiments we see no difference in terms of performance. However, since the former setting allows for the use of DQN \cite{mnih2013dqn} instead of DDPG \cite{lillicrap2015ddpg}, we prefer the discrete action space for simplicity and faster convergence. Both the MVProp (Equation \ref{eq:mvprop_main}) and Gaussian likelihood (Equation \ref{eq:gaussian_likelihood}) operations are differentiable. Hence, MVProp and the attention model parameters are trained by minimizing the standard mean squared Bellman error objective as defined in Equation \ref{eq:q_bellman}.

\subsection{Control Layer}
\label{sec4_controllayer}
The control layer learns a goal-conditioned control policy. Unlike the planning layer, it has access to the agent's internal state $s_{\text{internal}}$, including joint positions and velocities. In the control tasks we consider, the agent has to learn a policy to reach a certain goal position, e.g., reach a target position in a navigation domain. The agent-centric goal is provided by the planning layer. We use the hindsight techniques (cf. Section \ref{sec3_hac}) so that the control policy receives rewards even in failure cases. All policies in our hierarchy are trained jointly. We use DDPG \cite{lillicrap2015ddpg} (Equations \ref{eq:q_bellman}-\ref{eq:pi}) to train the control layer and DQN \cite{mnih2013dqn} for the planning layer.
\newlength\lengtha \setlength\lengtha{4mm} 
\vspace{-0.3cm}
\begin{figure}[h]%
    \centering
    \subfloat[Maze navigation]{{
    \includegraphics[width=0.30\columnwidth]{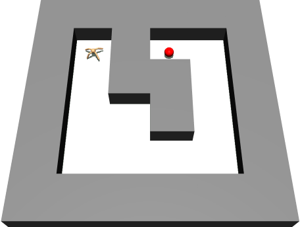} }}%
    \subfloat[\centering Manipulation]{{
    \includegraphics[width=0.27\columnwidth]{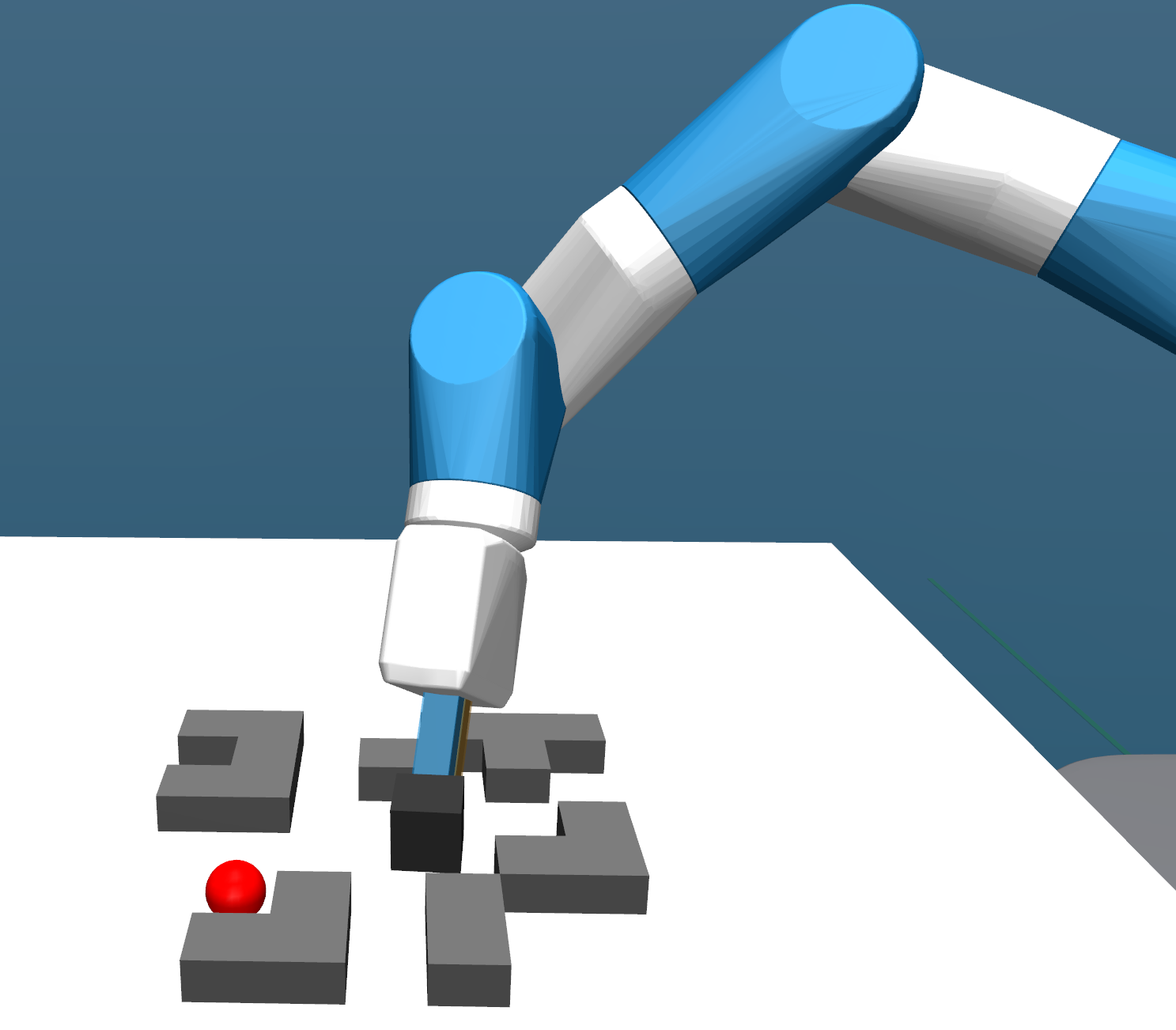} }}%
    \subfloat[3D Reacher]{
    {\includegraphics[width=0.37\columnwidth]{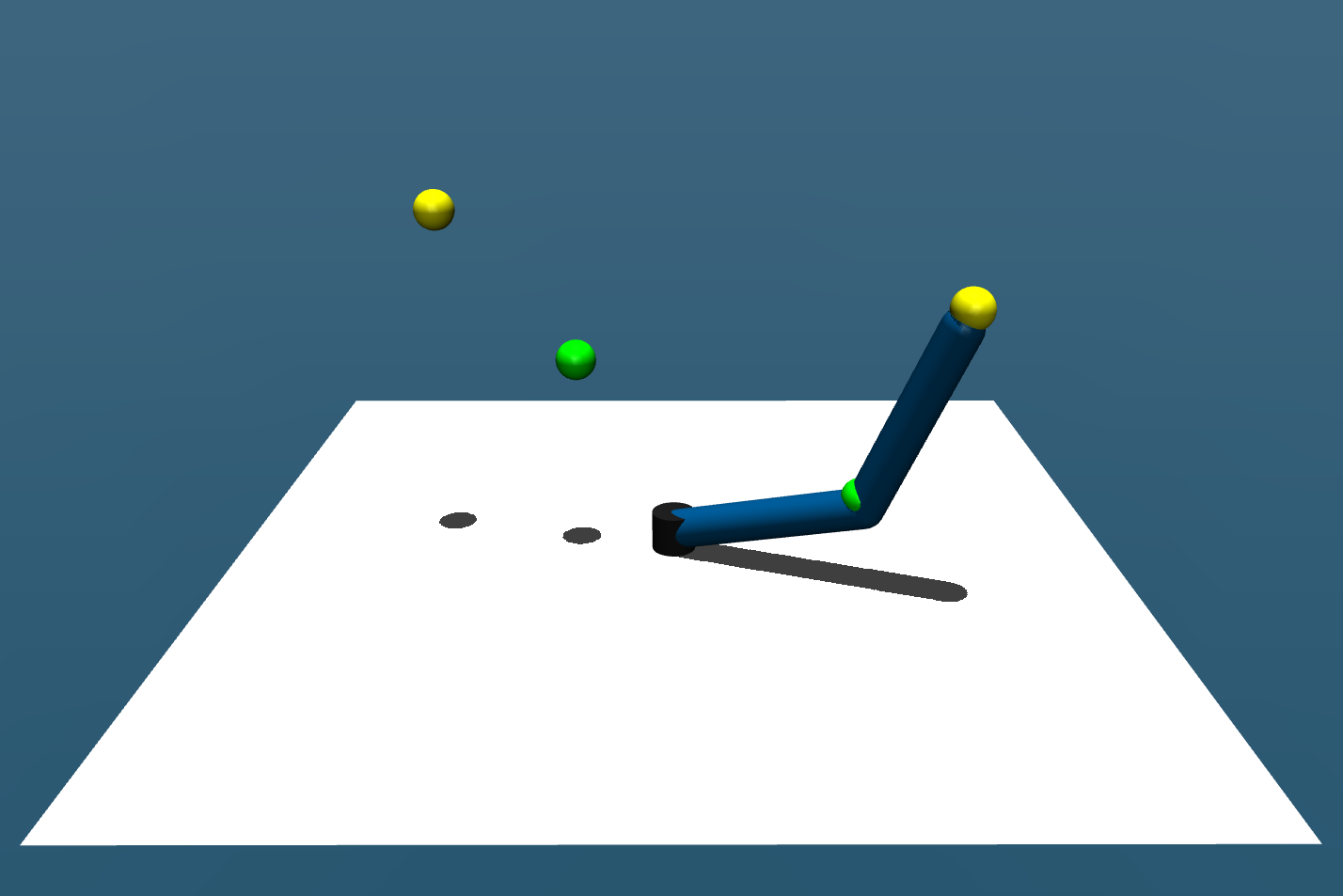} }}%
    \caption{a) The complex navigation training environment from Section \ref{subsec_mazenav_complex}. The red sphere indicates the goal. b) The robot manipulation task from Section \ref{sec:domain_transfer}. c) The 3D reacher task presented in Section \ref{sec:ext_epx}. The green and yellow spheres indicate the goals for the robot's elbow and tip.}%
    \label{fig:experiment_envs}%
    \vspace{-0.4cm}
\end{figure}

\begin{table*}[t]
    \vspace{0.2cm}
    \centering
    \caption{Success rates of the Maze navigation experiments.}
    \vspace{-0.2cm}
    \resizebox{0.8\linewidth}{!}{
    \begin{tabular}{@{}c|cccc|ccccc@{}}
        \hline
        \multicolumn{1}{}{} &  \multicolumn{4}{c}{Simple Maze} & \multicolumn{5}{c}{Complex Maze} \\ \hline
        Method & Training     & Backward    & Flipped  & Rotated & Training & Random  & Backward    & Flipped & Rotated \\ \hline
        HAC          & $82 \pm 16$ & $0 \pm 0$ & $0 \pm 0$ & $0 \pm 0$ & $0 \pm 0$ & $0 \pm 0$ & $0 \pm 0$ & $0 \pm 0$ & $0 \pm 0$ \\
        HIRO         & $91 \pm 2$  & $0 \pm 0$   & $0 \pm 0$ & $0 \pm 0$ & $68 \pm 8$ & $36 \pm 5$  & $0 \pm 0$ & $0 \pm 0$ & $0 \pm 0$ \\
        HIRO-LR         & $ 83 \pm 8 $  & $0 \pm 0$   & $0 \pm 0$ & $0 \pm 0$ &  $20 \pm 21$  & $15 \pm 7$   & $0 \pm 0$  & $0 \pm 0$ & $0 \pm 0$ \\
        RRT+LL      & $25 \pm 13 $  & $22 \pm 12$   & $29 \pm 10$ & $24 \pm 7$ & $13 \pm 9$ & $48 \pm 3$ & $7 \pm 4$ & $5 \pm 5$ & $8 \pm 8$\\
        RRT-S+LL      & $47 \pm 16 $  & $37 \pm 18$   & $47 \pm 17$ & $46 \pm 6$ & $32 \pm 13$ & $62 \pm 5$ & $20 \pm 12$ & $11 \pm 10$ & $19 \pm 17$\\
        RRT-S+LL*    & $77 \pm 7$ & $74 \pm 9$ & $61 \pm 12$ & $79 \pm 5$  & $69 \pm 5$  & $ 79 \pm 6$ & $54 \pm 13$ & $ 63 \pm 11$ & $59 \pm 10$\\
        HiDe         &$\mathbf{94 \pm 2}$  & $\mathbf{85 \pm 9}$ & $\mathbf{93 \pm 2}$ & $\mathbf{84 \pm 8}$ & $\mathbf{87 \pm 2}$ & $\mathbf{85 \pm 3}$ & $\mathbf{79 \pm 8}$ & $\mathbf{79 \pm 12}$ & $\mathbf{78 \pm 17}$ \\ \hline
    \end{tabular}}
    \label{tab:exp1}
    \vspace{-0.4cm}
\end{table*}

\begin{figure*}[t]%
    \centering
    \hspace{-0.7cm}
    \subfloat[Simple Maze Navigation]{{
    \includegraphics[width=0.9\columnwidth]{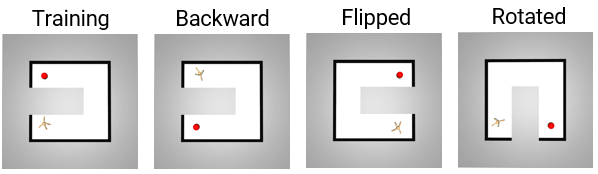} }}%
    \qquad
    \hspace{-0.6cm}
    \subfloat[Complex Maze Navigation]{
    {
    \includegraphics[width=0.57\columnwidth]{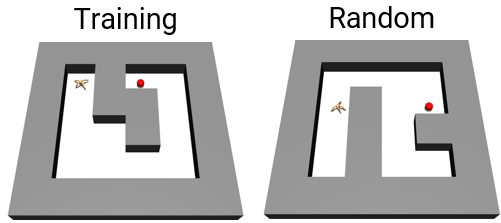} }}%
    \caption{Maze environments for the tasks reported in a)  Section \ref{sec5_exp1} and \ref{sec:ablation_study}, b) Section \ref{subsec_mazenav_complex}. The red sphere indicates the goal. Agents are always trained in the \textit{Training} environment and tested in the \textit{Backward}, \textit{Flipped}, and \textit{Random} environments.}%
    \label{fig:navigation_envs}%
    \vspace{-0.5cm}
\end{figure*}

\section{Experiments}
\label{sec5_experiments}
We evaluate our method on a series of continuous control tasks which are implemented in Mujoco \cite{todorov2012mujoco}. First, we compare to various baseline methods in navigation tasks (see Fig. \ref{fig:navigation_envs}a) in Section \ref{sec5_exp1} and show in Section \ref{subsec_mazenav_complex} that HiDe can scale beyond 3x larger environments (see Fig. \ref{fig:navigation_envs}b). In Section \ref{sec:ablation_study}, we provide an ablation study for our design choices. Section \ref{subsec_transfer_policies} demonstrates that our approach indeed leads to functional decomposition by transferring layers across agents and domains (see Fig. \ref{fig:experiment_envs}b). We show in Section \ref{sec:ext_epx} that HiDe can be extended to planning in 3D and use non-image based priors, such as a joint map (see Fig. \ref{fig:experiment_envs}c). The code with pretrained models along with videos of the experiments is made publicly available\footnote{Videos and code available at \small{{https://sites.google.com/view/hide-rl}}\label{video_ref}}.

\subsection{Maze Navigation}
\label{sec:maze_nav}

We introduce the following task configurations:\\
\textbf{Maze Training}  The training environment, where the task is to reach a goal from a fixed start position. \\
\textbf{Maze Backward} The training environment with swapped start and goal positions. \\
\textbf{Maze Flipped} The mirrored training environment.\\
\textbf{Maze Rotated} The training environment rotated by 90°.\\
\textbf{Maze Random} A set of 500 randomly generated mazes with random start and goal positions.

We always train in the Maze Training environment.  The rewards in the experiments are sparse, i.e., $0$ for reaching the goal and ${-1}$ otherwise (except for HIRO/HIRO-LR with an L2-shaped reward). We test the agents on the above task configurations. We intend to answer the following questions:

\begin{enumerate}
    \item Can our method generalize to unseen test cases and environment layouts?
    \item Can we scale to larger environments with more complex layouts (see Fig. \ref{fig:navigation_envs}b)?
\end{enumerate}

We compare our method to state-of-the-art HRL approaches including HIRO \cite{nachum2018data}, HIRO-LR \cite{nachum2018nearoptimal}, HAC \cite{levy2018hierarchical}, and a set of more conventional navigation baselines using RRT \cite{rrt}. HIRO-LR is the closest related work, since it receives an occupancy grid and is a fully learned hierarchy.

Our preliminary experiments have shown that HAC and HIRO cannot solve the task when provided with an occupancy grid, likely due to the  increase of the state space by factor of 14. We therefore only show results of HAC and HIRO where they are able to solve the training task, i.e., without accessing an occupancy grid.  To compare against a baseline with complete separation, we introduce RRT+LL. We train a goal-conditioned control policy with RL in an empty environment and attach an RRT planner \cite{rrt}, which finds a path from top-down views via tree-search and does not require training. We also introduce RRT-S+LL and RRT-S+LL*, where a hand-tuned safety margin is added to the planner such that it does not select subgoals close to the walls. In RRT-S+LL*, we add more noise to the starting states, which leads to more robust control policies.
All methods were trained for 2.5 million steps (simple maze, Fig. \ref{fig:navigation_envs}a) and 10 million steps (complex maze, Fig. \ref{fig:navigation_envs}b).

\subsubsection{Simple Maze Navigation}
\label{sec5_exp1}

\begin{figure*}[t]%
    \centering
    \subfloat[Simple Maze Navigation Training Curves]{{
    \includegraphics[width=0.7\columnwidth]{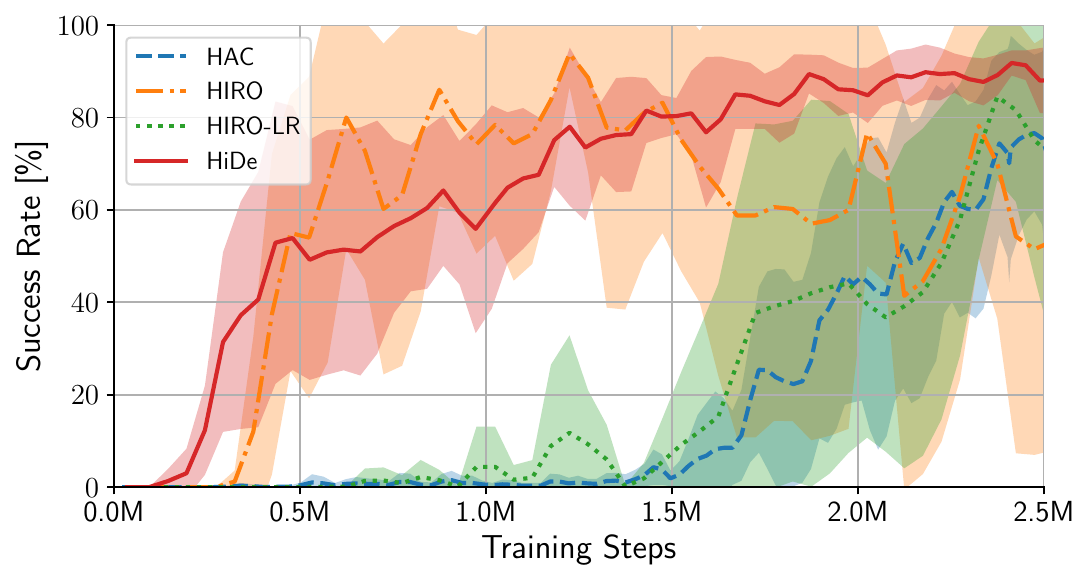} }}
    \qquad
    \subfloat[Complex Maze Navigation Training Curves]{{
    \includegraphics[width=0.7\columnwidth]{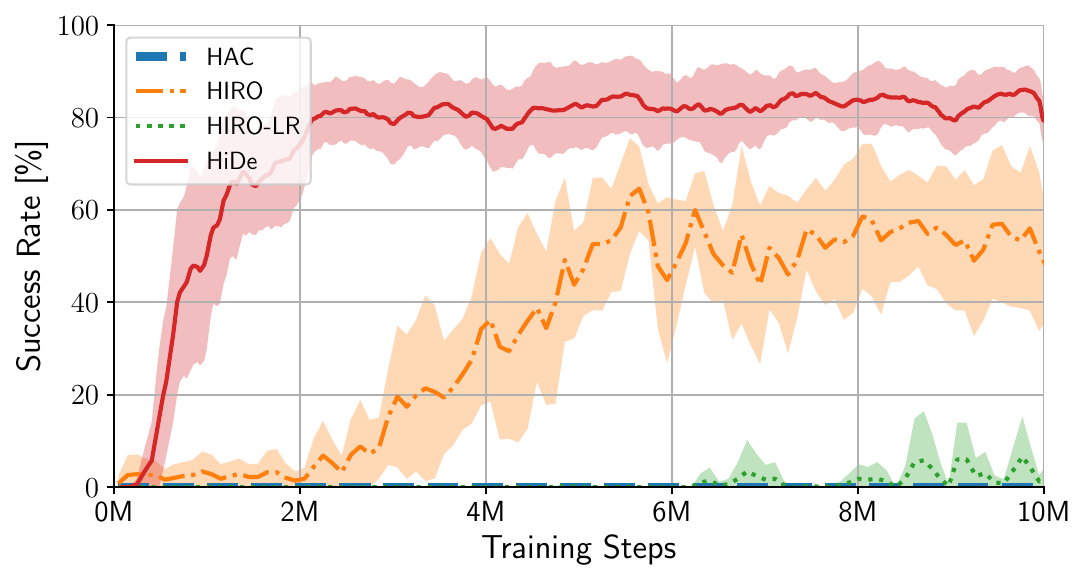} }}
    \vspace{-0.1cm}
    \caption{Learning curves with success rates for the training tasks averaged over 5 seeds for a) the simple maze experiment from Section \ref{sec5_exp1}, b) the complex maze experiment from Section \ref{subsec_mazenav_complex}. HiDe matches convergence properties of HIRO in the simple maze (left), albeit having a much larger state space in the planning layer. In the more complex maze (right), HiDe shows similar convergence, while the convergence of HIRO slows.}%
    \label{fig:curves}%
    \vspace{-0.5cm}
\end{figure*}

Table \ref{tab:exp1} (left) summarizes the results for the simple maze tasks shown in Fig. \ref{fig:navigation_envs}a. All HRL models successfully learned the training task (Forward). The models' generalization abilities are evaluated in the unseen Maze Backward, Flipped and Rotated tasks. While HIRO, HIRO-LR and HAC manage to solve the training environment with success rates between 91\% and 82\%, they overfit to the training task, indicated by the 0\% success rates in the unseen test cases. HIRO-LR, which uses the top-down view implicitly to learn a goal space representation, also fails. 
For the RRT+LL baseline, the success rate stays between 22\% and 29\% for all tasks, with failures arising mostly due to the agent falling over. The success rate slightly improves in RRT-S+LL due to the added safety margin. Adding more noise to the starting states helps to further increase the performance (RRT-S+LL*). However, although the planner can generalize to different tasks, it cannot learn to cooperate with the low-level control as in our hierarchy. Our method is able to achieve 93\% and 85\% success rates in the generalization tasks \emph{without} retraining. We argue that this is mainly due to the strict separation of concerns, which allows the integration of task-relevant priors, in combination with HiDe's efficient planner and the emerging curriuculum for the control policy. 

\subsubsection{Complex Maze Navigation}
\label{subsec_mazenav_complex}

In this experiment, we evaluate how well the methods scale to larger environments with longer horizons. Thus, we train an ant agent in a more complex environment layout (cf. Fig. \ref{fig:navigation_envs}b), i.e., we increase the size of the environment by roughly 50\% and add more obstacles, thereby also increasing the distance to the final reward. The results are reported in Table \ref{tab:exp1} (right). HAC fails to learn the training task, while HIRO and HIRO-LR reach success rates of 68\% and 20\%, respectively. Hence, there is a significant performance drop for both methods if the state-space increases. The best performing RRT baseline, RRT-S+LL* reaches success rates between 54\% and 69\%, except for the Maze Random task. Note that HiDe does not have access to such a safety margin, yet learns to avoid the walls. The higher success rates in Maze Random compared to the other test cases is due to the randomization of both the environment layout as well as the start and goal position, which can result in short trajectories without obstacles. 
Contrary to the baselines, our model's performance decreases only slightly in the training task compared to the simple maze and also generalizes to all of the unseen test cases. The decrease in performance is due to the increased difficulty of the task. We analyze the convergence behavior between the different methods in Fig. \ref{fig:curves}. We find that in the simple maze case HIRO is competitive with HiDe, due to its reduced state space (no grid) and access to a shaped L2-based reward, but convergence slows with an increase in complexity. Contrarily, HiDe shows similar convergence behavior in both experiments. To push the limits of our method, we gradually increase the environment size and observe that only at a 300\% increase, the performance drops to around 50\%. These results indicate that task-relevant information and efficient methods to leverage it are essential components to scale to larger environments. Most remaining failure cases for HiDe arise if the agent flips over.

\subsubsection{Efficiency of Policies}
\label{subsec_optimality}
\begin{table}[t]
    \caption{Efficiency Comparison. We measure the number of control actions required to reach the target in successful trials.}
    \centering
    \resizebox{0.6\linewidth}{!}{
    \begin{tabular}{@{}c|c|c@{}}
    \hline
    Methods & Simple Maze     & Complex Maze   \\ \hline
    HIRO    & $184 \pm 100 $  & $323 \pm 28$  \\
    RRT+LL    & $195 \pm 15 $  & $354 \pm 42$ \\ 
    RRT-S+LL*    & $157 \pm 19 $  & $247 \pm 5$  \\
    HiDe         & $169 \pm 7$  & $250 \pm 32$ \\\hline
    HiDe-S   & $\mathbf{141 \pm 13}$ & $\mathbf{209 \pm 5}$ \\ \hline
    \end{tabular}}
    \vspace{-0.4cm}
    \label{tab:optimal}
\end{table} 

To assess the efficiency of the policies, we evaluate the average number of steps it takes an agent to reach a target in successful trials. We compare against HIRO as the best performing learning-based method and the different RRT variants in both the simple and complex maze. The results are summarized in Table \ref{tab:optimal}. We find that adding safety margins to HiDe (HiDe-S) generates the best results and use it as an upper baseline. HIRO takes inefficient routes and tends to cut corners, likely due to the L2-based reward it relies on. The naive RRT+LL baseline shows similar behavior. Adding a safety margin and using more robust low-level policies (RRT-S+LL*) mitigates this and increases the performance. HiDe shows similar scores as the RRT-S+LL* baseline, although it does not have access to the safety margin, indicating that it implictly learns to avoid goals close to the walls. The performance gap between HiDe and HiDe-S implies that HiDe may still sometimes provide goals too close to the walls, which could be improved upon, e.g., by adding an additional penalty to the planner for subgoals where the agents comes in contact with a wall.

\begin{table}[t]
    \caption{Ablation study in the simple maze navigation environments.}
    \centering
    \resizebox{0.6\linewidth}{!}{
    \begin{tabular}{@{}c|c|c|c@{}}
    \hline
    Methods & Training     & Backward    & Flipped     \\ \hline
    HiDe-A    & $88 \pm 2 $  & $17 \pm 15$   & $36 \pm 16$   \\ 
    HiDe-3x3    & $46 \pm 32 $  & $2 \pm 3$   & $31\pm 28$   \\
    HiDe-5x5    & $92 \pm 4 $  & $41 \pm 35$   & $82\pm 18$   \\
    HiDe-9x9    & $93 \pm 4 $  & $16 \pm 27$   & $79\pm 7$   \\
    HiDe-RRT         & $77 \pm 9$  & $53 \pm 13$ & $72 \pm 6$ \\
    HiDe         & $\mathbf{94 \pm 2}$  & $\mathbf{85 \pm 9}$ & $\mathbf{93 \pm 2}$ \\ \hline
    \end{tabular}}
    \vspace{-0.17cm}
    \label{tab:ablation}
\end{table} 

\subsection{Ablation Study}
\label{sec:ablation_study}

\begin{figure}[t]
\centering
\includegraphics[width=0.5\columnwidth, trim={0 245 390 0}]{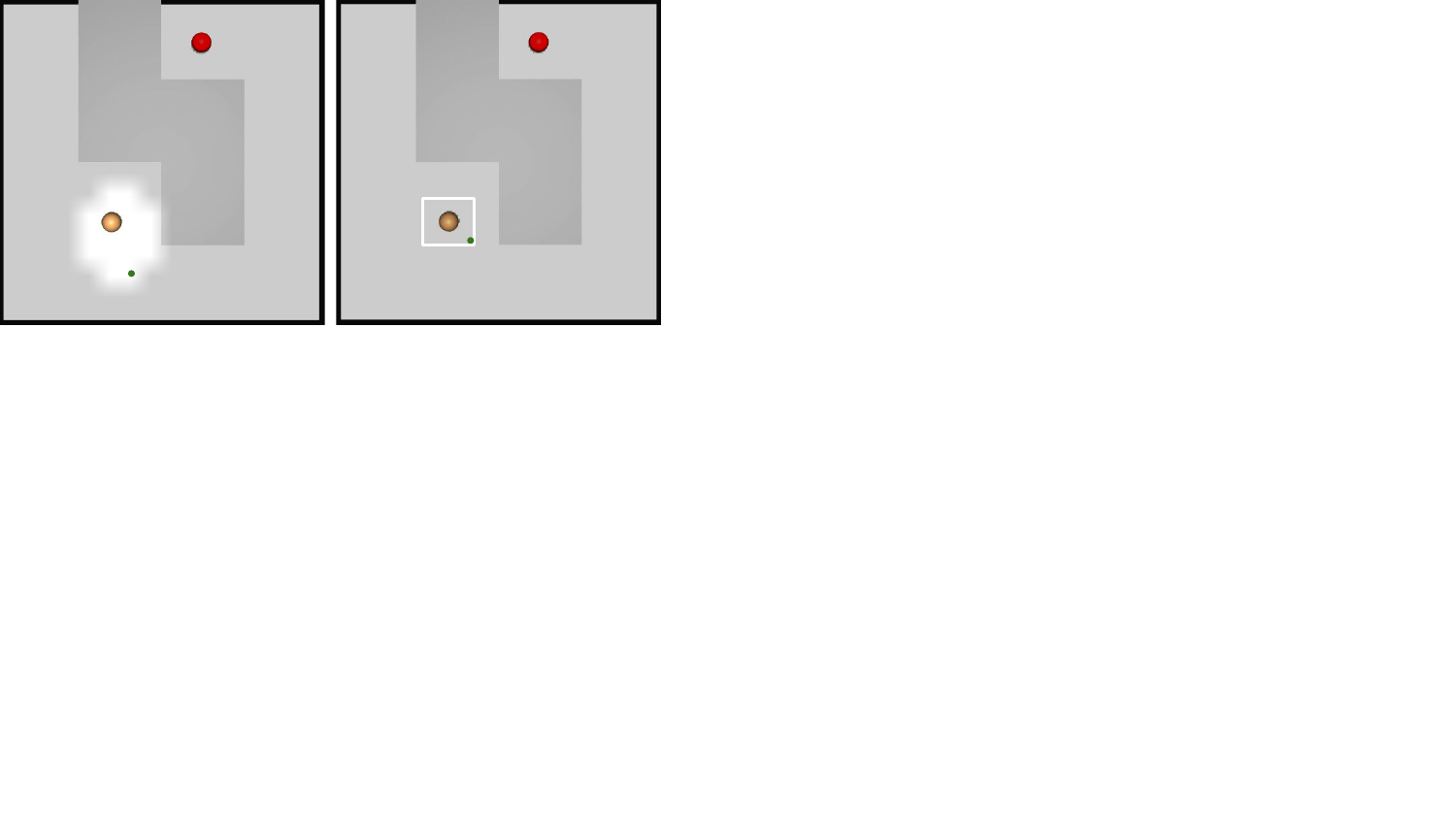}
\vspace{-0.1cm}
\caption{A visual comparison of \textit{(left)} our dynamic attention window with a \textit{(right)} fixed neighborhood. The green dot corresponds to the selected subgoal.  Notice how our window is shaped so that it avoids the wall.}
\label{fig:dynamic_static}
\vspace{-0.4cm}
\end{figure}

To support the claim that our architectural design choices support the generalization and scaling capabilities, we analyze empirical results of different variants of our method. To show the benefits of relative positions, we compare HiDe against a variant with absolute positions, dubbed HiDe-A. Unlike the case of relative positions, the policy needs to learn all values within the range of the environment dimensions in this setting. Second, we run an ablation study for HiDe with a fixed window size, i.e., we train and evaluate an ant agent on window sizes $3\times3$, $5\times5$, and $9\times9$. Lastly, we compare to a variant where HiDe's decoupled state-space structure is used for training, but the RL-based planner is replaced with an RRT planner. As indicated in Table \ref{tab:ablation}, HiDe-A is competitive in the training task, but fails to match the generalization performance of HiDe. showing that relative positions are crucial for generalization. The learned attention window (HiDe) achieves better or comparable performance than the fixed window variants. We also observe qualitatively that behavior around corners is better compared to the fixed window, as it can learn to avoid corners (see Fig. \ref{fig:dynamic_static}). 
Moreover, it eliminates the need for tuning the window size per agent and environment. HiDe-RRT performs significantly worse in all tasks, showing that our learned planner outperforms training with RRT and hence creates a beneficial curriculum for the control policy.

\subsection{Transfer of Policies}
\label{subsec_transfer_policies}

We argue that a key to transferability and  generalization behavior in hierarchical RL lies in enforcing a separation of concerns across different layers. we therefore perform a set of experiments to demonstrate transfer behavior.

\subsubsection{Agent Transfer}
\label{sec:exp_agent_transfer}


For this experiment, we train different control agents with HiDe in the complex maze environment (see Fig. \ref{fig:navigation_envs}b). We then transfer the planning layer of one agent to another agent, e.g., we replace the planning layer of a complex ant agent by the planning layer trained on a simple $2$ DoF ball agent and vice-versa. We compare against a recent method that allows for agent transfer \cite{abbeel_scoop} and uses a discriminator to improve the domain shift. We apply the zero-shot transfer and low-level imitation versions of \cite{abbeel_scoop}, which we call "HMT Zero-Shot" and "HMT Discrim.", and omit the variant with fine-tuning of the high-level policies.

The results are illustrated in Table \ref{tab:agent_transfer}. For the transfer of the ball's planner to the ant (Ball PL $\rightarrow$ Ant CL), we observe that transfer with HiDe outperforms the HMT baselines and only leads to a marginal decrease in performance compared to the results reported in Table \ref{tab:exp1}. Most failure cases arise at corners, where the ball's planner tries to use a path close to the walls. Contrarily, the ant's planner is more conservative as subgoals close to the wall may lead to overturning.  
When transferring the ant's planner to the ball agent (Ant PL $\rightarrow$ Ball CL), there is no decrease in performance for HiDe. On the other hand, the HMT variants only reach low success rates of 12-20\% in the simple maze and fail in the complex maze, because training the ant planner seldomly succeeds.
We hereby show that HiDe can achieve transfer of layers between agents and outperform a state-of-the-art method \cite{abbeel_scoop}. To further demonstrate our method's transfer capabilities, we train a humanoid agent (17 DoF) in an empty environment. We then use the planning layer from ball agents and connect it as is with the control layer of the humanoid. We report results on two different sets of 500 random layouts, Complex Random as the same set from Section V-A and Simple Random with less clutter. The success rates of 37\% and 19\% (cf. Table \ref{tab:agent_transfer}) indicate the fragility of the humanoid control policies and leave room for improvement. We will release our test environments to foster further research.

\begin{table}[t]
    
    \caption{Transfer Comparison Results.}
    \centering
    \resizebox{\linewidth}{!}{
    \begin{tabular}{@{}c|c|c|c@{}}
        \hline
        \multirow{5}{*}{Ball PL $\rightarrow$ Ant CL} &
        Method & Simple Maze & Complex Maze \\ \hline
        & HMT Zero-Shot \cite{abbeel_scoop}    & $42 \pm 18$ & $16 \pm 14$ \\
        & HMT Discrim. \cite{abbeel_scoop}     & $ 54 \pm 14$ & $20 \pm 18$\\
        & HiDe         & $\mathbf{84 \pm 11}$ & $\mathbf{72 \pm 2}$ \\\hline
        \multirow{3}{*}{Ant PL $\rightarrow$ Ball CL} 
        & HMT Zero-Shot \cite{abbeel_scoop}    & $12 \pm 26$ & $0 \pm 0$\\
        & HMT Discrim. \cite{abbeel_scoop}     & $20 \pm 45$ & $0 \pm 0$ \\
        & HiDe         & $\mathbf{100 \pm 0}$ & $\mathbf{100 \pm 0}$\\\hline
        &  & Simple Random & Complex Random \\ \hline
        Ball PL $\rightarrow$ Humanoid CL & HiDe  & $\mathbf{37 \pm 3}$ & $\mathbf{19 \pm 4}$ \\\hline
        Ball PL $\rightarrow$ Robot Arm CL & HiDe  & $\mathbf{49 \pm 1}$ & $\mathbf{33 \pm 3}$ \\ \hline

    \end{tabular}}
    \label{tab:agent_transfer}
    \vspace{-0.2cm}
\end{table}

\subsubsection{Domain Transfer}
\label{sec:domain_transfer}
In this experiment, we demonstrate the capability of HiDe to transfer the planning layer from a simple ball agent, trained on a pure navigation task, to a robot manipulation agent (see Fig. \ref{fig:experiment_envs}b). To this end, we train a ball agent with HiDe. Moreover, we train a control policy for a robot manipulation task in the OpenAI Gym "Push" environment \cite{openai_gym}, which learns to move a cube to a relative position goal. Note that the manipulation task does not encounter any obstacles during training. To attain the compound agent, we attach the planning layer of the ball agent to the manipulation policy (cf. Fig. \ref{fig:teaser}a). The planning layer has access to the environment layout and the cube's position, which is a common assumption in robot manipulation tasks. For testing, we generate 2 sets of 500 random environment layouts, one set with few obstacles (Simple Random) and a more cluttered set (Complex Random). As in the navigation experiments in Section \ref{sec5_exp1}, state-of-the-art methods are able to solve these tasks when trained on a single, simple environment layout. However, they do not generalize to other layouts without retraining. In contrast, our evaluation of the compound HiDe agent on unseen testing layouts shows a success rate of 49\% and 33\% (cf. bottom row of Table \ref{tab:agent_transfer}) in the two random test sets. Most failure cases arise when the robot's end-effector gets stuck at an obstacle. Thus, our modular approach can achieve domain transfer and generalize to different environments, although leaving room for improvement.

\subsection{Representation of Priors and 3D-Planning}
\label{sec:ext_epx}

While the majority of our experiments use 2D-grids for planning, we provide a proof-of-concept that our method i) can be extended to planning in 3D ii) works with non-top-down view sources of information. To this end, we add a 3DoF robotic arm that has to reach goals in 3D configuration space (see Fig. \ref{fig:experiment_envs}c). Instead of a top-down view, we project the robot's joint angles onto a 3D-map which is used as input to our planner. The planner finds subgoals that bring the robot into the target pose. The low-level policy learns a controller that applies motor torques in order to reach the given subgoal poses of the planner. We train the 3D variant of our planning layer (see Section \ref{sec4_planninglayer}) and 3D CNNs to compute the value map. Our method can successfully solve the task (see supplementary video). If the planning space exceeds 3 dimensions, a mapping from higher dimensional representation space to a latent space could be a solution. 
\section{Discussion \& Conclusion}
\label{sec:conclusion}

In this paper, we introduce a novel HRL architecture that can solve long-horizon, continuous control tasks with sparse rewards that require planning. The architecture, which is trained end-to-end, consists of an RL-based planning layer which learns an explicit value map and is connected with a low-level control layer. Our method is able to generalize to unseen settings and environments. Furthermore, we show that transfer of planners between different agents can be achieved, enabling us to move a planner trained on a simple agent to a more complex agent, such as a humanoid. 
The key insight lies in a strict separation of concerns with task-relevant priors that allow for efficient planning and in consequence better generalization. 
Several issues remain. In the simulation, we operate under the ideal assumption that an occupancy grid is available, which may not be the case in real-world settings. A potential solution to the problem could be to first use SLAM with a simple robot to obtain an occupancy grid of the environment and subsequently apply HiDe with an added sensing module to estimate the position of the agent in the environment.
Furthermore, when transferring the learned attention mask, the planner does not take into account the capabilities of the control agent. This leads to a decrease in performance (cf. Section V-C), which could be addressed by fine-tuning of the higher layer policies, such as suggested in \cite{abbeel_scoop}. Other Interesting directions for future work include experiments on a real-world robot or multi-agent settings. 

\section{Acknowledgements}

This project has received funding from the European Research Council (ERC) under the European Union’s Horizon 2020 research and innovation programme grant agreement No. StG-2016-717054.


\bibliographystyle{IEEEtran}
\bibliography{refs}

\end{document}